\documentclass[runningheads]{llncs}

\usepackage{bm,booktabs,placeins,amsmath}
\usepackage{hyperref}
\usepackage{cleveref}
\usepackage{fancyhdr}

\pdfoutput=1

\usepackage{graphicx}
\begin{document}


\title{Robust Angular Local Descriptor Learning\thanks{Supported by grant  Pfizer and organization by SAP SE and CNRS INS2IJCJC-INVISANA.}} 
\titlerunning{RAL-Net} 


\author{Yanwu Xu\inst{1,3} \and
Mingming Gong\inst{1} \and
Tongliang Liu\inst{2}\and
Kayhan Batmanghelich\inst{1}\and
Chaohui Wang\inst{3}}
%

\authorrunning{Yanwu et al.} 


\institute{University of Pittsburgh,  4200 Fifth Avenue
Pittsburgh, PA 15260, USA \email{\{yanwuxu,mig73,kayhan\}@pitt.edu}\\
\and
The University of Sydney, Camperdown NSW 2006, Australia \email{tongliang.liu@sydney.edu.au}\\
\and
Universit\'{e} Paris-Est, LIGM (UMR 8049), CNRS, ENPC, ESIEE Paris, UPEM, Marne-la-Vall\'{e}e, France \\ \email{chaohui.wang@u-pem.fr}}

\maketitle\thispagestyle{fancy}

\begin{abstract}
In recent years, the learned local descriptors have outperformed handcrafted ones by a large margin, due to the powerful deep convolutional neural network architectures such as L2-Net \cite{8100132} and triplet based metric learning \cite{hard}. However, there are two problems in the current methods, which hinders the overall performance. Firstly, the widely-used margin loss is sensitive to incorrect correspondences, which are prevalent in the existing local descriptor learning datasets. Second, the L2 distance ignores the fact that the feature vectors have been normalized to unit norm. To tackle these two problems and further boost the performance, we propose a robust angular loss which 1) uses cosine similarity instead of L2 distance to compare descriptors and 2) relies on a robust loss  function that gives smaller penalty to triplets with negative relative similarity. The resulting descriptor shows robustness on different datasets, reaching the state-of-the-art result on Brown dataset , as well as demonstrating excellent generalization ability on the Hpatches dataset and a Wide Baseline Stereo dataset. Our codes is released in github\footnote{\url{https://github.com/xuyanwu/RAL-Net}}

\end{abstract}
\section{Introduction}

Finding correspondences between local patches across images is an important component in many computer vision tasks, such as image matching \cite{NIPS2016_6487}, image retrieval \cite{4270197} and object recognition \cite{790410}. Since the seminal paper introducing SIFT \cite{Lowe2004}, local patches have been encoded into representative vectors, called descriptors, which are designed to be invariant/robust to various geometric and photometric changes such as scale change, viewpoint change, and illumination change.

Given the success of deep learning, hand-crafted descriptors such as SIFT have been outperformed by learned ones \cite{2014arXiv1405.5769F,7410379,7298948}. Different from the hand-crafted descriptors which extract low-level features such as gradients, the learned descriptors learn a convolutional neural network (CNN) from raw patches with ground-truth correspondences. These descriptor learningnetworks are trained by metric learning losses and can be divided into two cat-egories by whether there are learnable distance comparison layers in the network. The networks with distance comparison layers output distances directly without explicit descriptors \cite{7298948,2015arXiv150403641Z,6718113}. This type of networks showed promising performance in patch verification but cannot be combined with nearest neighbor search. Recently, networks without similarity comparison layers achieved better performances due to more advanced network architectures such as L2Net \cite{8100132} and training techniques such as triplet loss with hard negative mining \cite{hard}. These networks output descriptors which can be compared using simple L2 distance and be matched using fast approximate nearest neighbor search algorithms like Kd-tree \cite{Bentley:1975:MBS:361002.361007}.

\vspace{-0.3cm}
\begin{figure}[htb]
\begin{minipage}[b]{1.0\linewidth}
  \centering
  \centerline{\includegraphics[width=0.8\textwidth]{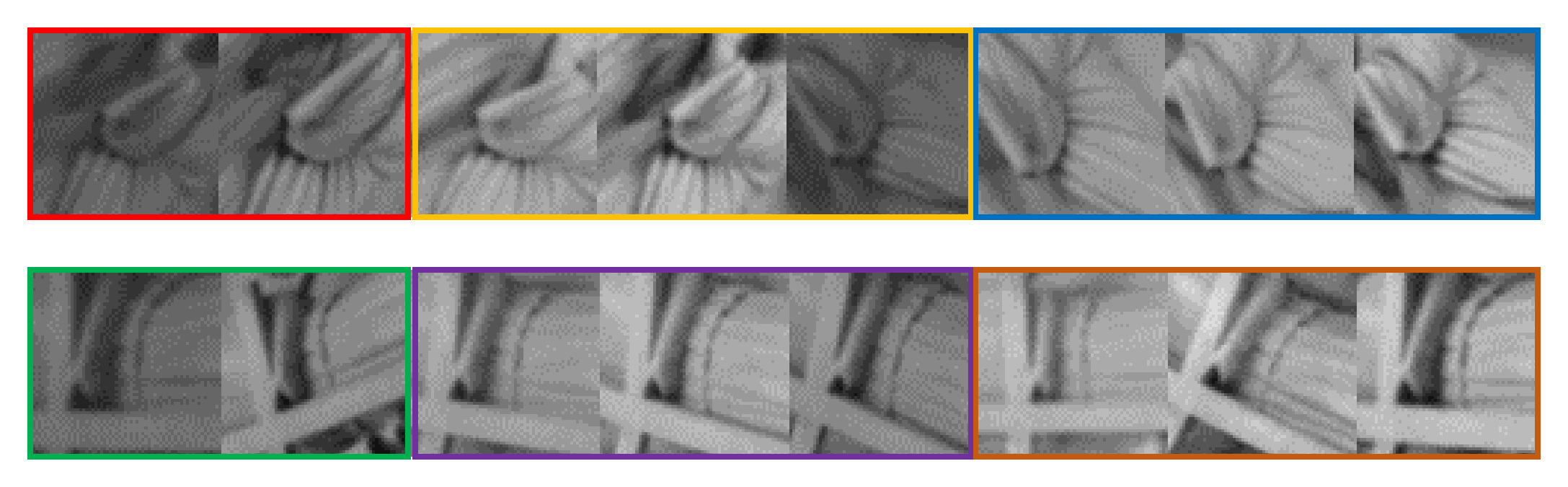}}
  \centerline{}\medskip
  \vspace{-1cm}
\end{minipage}
\caption{Examples of false labeled patches, the patches sharing same label of 3D view point are marked by same color box,and different color boxes come from different 3D view point.}
\label{false}
\vspace{-0.3cm}
\end{figure}

In the descriptor learning networks, the metric learning loss function and the distance/similarity measure between descriptors are two essential components. State-of-the-art methods usually adopt margin-based losses such as the hinge loss \cite{hard} to train the descriptor learning networks. Because the number of negative pairs is huge, batch hard negative mining is usually applied to stabilize the training process as well as reduce the computational load \cite{hard,7298682}. However,
the current triplet losses are not robust to the incorrect correspondences (outliers) in the training data, as shown in Fig ~\ref{false}. The patches at different locations (negative pairs) can exhibit strong similarities and the patches at the same location (positive pairs) can be very different due to local distortion or corruptions. Additionally, since the local descriptors are normalized to unit norm before comparison, L2 distance is no longer an appropriate distance measure to compare
descriptors.

To target the these two problems, we propose a robust angular loss to train
the descriptor learning networks which is called RAL-Net. Instead of using the
hinge loss as done in \cite{hard}, we propose a robust loss function which gives bounded
penalty to the triplets with incorrect correspondences. In addition, we propose to
utilize cosine similarity to compare two descriptors, which is more appropriate
to compare unit-norm vectors. We train our descriptor on the Brown dataset
\cite{5432199} and obtain state-of-the-art results using the same training strategy as \cite{hard}.
Moreover, our descriptor performs much better than \cite{hard} when the sample size
and batch size is small, which further verifies the effectiveness of the proposed
method.


\section{Related works}

Recent work on local descriptor designing has gone through a 
huge change from conventional hand-crafted descriptors to learning-based approaches, which ranges from SIFT \cite{Lowe2004} and DAISY \cite{daisy} to latest methods such as DeepCompare, MatchNet, and HardNet \cite{2014arXiv1405.5769F,7410379,7298948,hard}. As for deep learning-based descriptors, there are two study trends including CNN structure designing and negative sampling for embedding learning. 

Before CNN models being broadly applied, descriptors learning methods were limited to specific machine learning descriptors. Therefore, there were various kinds of methods inspired by different aspects. Principal Components Analysis (PCA) based SIFT (PCA-SIFT) \cite{Ke04pca-sift:a} leads to normalized gradient patch compared to SIFT histograms of gradients. \cite{5432199} proposed a filter with learned pooling and dimension reduction. Simonyan et al. \cite{6718113} studied convex sparse learning to learn pooling fields for descriptors. Aside from these descriptors, \cite{7298850} raised an online search method from a subset of tests which can increase inter-class variance and decrease intra-class variance. One thing these methods have in common is that they all rely on shallow learning architectures.

In the past few years, models based on CNN try to get better performance by designing various convolutional neural network architectures, e.g. \cite{7298948,2015arXiv150403641Z}. \cite{7298948} choose a two-branched network, a typical Siamese structure for feature extraction and three full connected layers for deep metric learning. \cite{2015arXiv150403641Z} explored further on Siamese network with two branches sharing no parameters and proposed a two-channel input structure which is stacked by center cropped patches and plain patches.

Recently methods focused more on loss function design because improving network structure can not give birth to significant improvement of descriptors as before. These works on learning embedding can be summarized as classification loss, contrastive loss, and triplet loss.  \cite{NIPS2014_5349,6909640} proved the validity classification loss for face recognition ans scene recognition. As the most common pairwise loss, contrastive loss \cite{1640964,DBLP:journals/corr/VariorHW16} aims at increase all of the similarity of positive pairs and push away the negative pairs until bigger than a variant margin. \cite{DBLP:journals/corr/LinMCVG15} proposed a restricting two sides margin for contrastive learning and this method not only requires distance between positive pairs above the margin but also limits distance between positive pairs under the second margin. Compared to contrastive loss, triplet loss cares about relative similarity between positive pairs and negative pairs rather than absolute value which consists of anchor positive $(a,p)$ and anchor negative $(a,n)$ with a shared point anchor $a$. This method can meet with most of tasks involving large scale embedding learning \cite{Chechik:2010:LSO:1756006.1756042}. However, there is still a great challenge in choosing hinge margin for triplet loss, as well as searching proper negative pairs fed into triplet loss. Distinguished from this embedding learning methods, Histogram loss \cite{NIPS2016_6464} uses a quadruplet based sampling strategy by estimating distribution of similarity between positive pairs and negative pairs.

From previous studies, the loss functions have their advantages and disadvantages in different learning tasks. A indispensable component of these methods is the negative sampling strategy. In this paper, our main aim is to improve the loss function to further improve the performance of local descriptor learning. Our method can possibly be applied to other related tasks such as face recognition and person re-identification.


\section{Proposed method}

In this section, we will discuss the form of our robust triplet loss which similar to \cite{Yu2010RelaxedCA} and simply introduction to network structure which is based on \cite{8100132}. In order to explain our loss function, we first review the general forms of triplet loss and contrastive loss then present our difference.

\subsection{loss function}
\label{ssec:subhead}
\begin{figure}[htb]
\begin{minipage}[b]{0.48\linewidth}
  \centerline{\includegraphics[width=1.3cm]{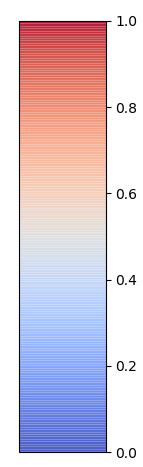}}
  \centerline{(a) Derivative colormap}\medskip
\end{minipage}
\hfill
\begin{minipage}[b]{0.48\linewidth}
  \centerline{\includegraphics[width=6cm]{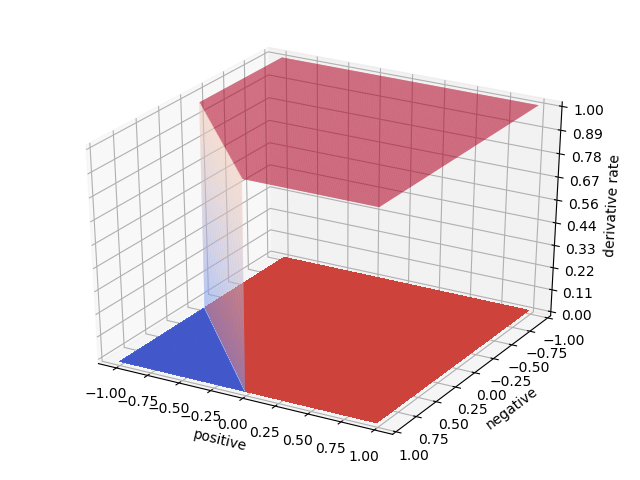}}
  \centerline{(b) Triplet loss}\medskip
\end{minipage}
\hfill
\begin{minipage}[b]{0.48\linewidth}
  \centerline{\includegraphics[width=6cm]{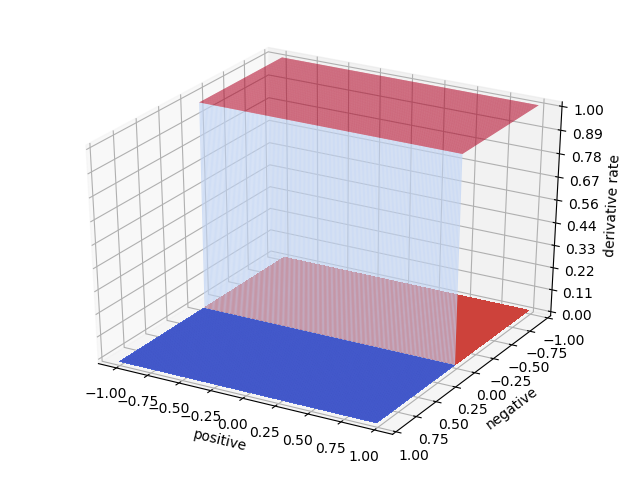}}
  \centerline{(c) Contrastive loss}\medskip
\end{minipage}
\hfill
\begin{minipage}[b]{0.48\linewidth}
  \centerline{\includegraphics[width=6cm]{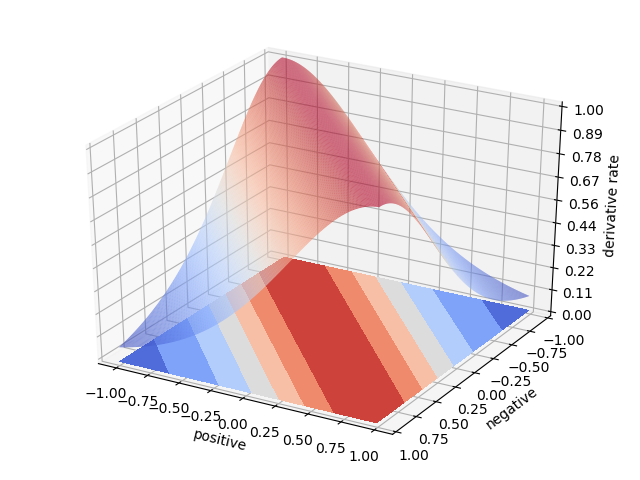}}
  \centerline{(d) Robust loss}\medskip
\end{minipage}
\caption{This figure depicts the derivation of three different loss function, (b)triplet loss,(c)contrastive loss,(d)robust loss. the cosine similarity of positive pairs and negative pairs represent x axis and y axis,where z axis denotes the absolute derivative values with respect to the change of x and y. The derivative value becomes larger when approaching red, vice versa when approaching blue.}
\label{derivative}
\vspace{-0.3cm}
\end{figure}

\subsection{The triplet loss}
Triplet loss has been successfully applied to many tasks, such as image matching, image retrieval, and face identification. The idea is to make positive pairs closer and keep relative negative pairs away from the positive. The very common expression of triplet loss is formulated as follows:
\begin{equation}
L_{triplet}=\frac{1}{N}\sum_{i,j,k}[s(a_{i},n_{k}) - s(a_{i},p_{j})+m]_{+}
\end{equation}
where $a, p$ and $n$ represent anchor, positive and, negative of triplet tuple and operator $[l]_{+}$ means $max(l,0)$ and function $s(x,y)$ represent similarity score between two featrues. Due to the large amounts of combination among $a, p$ and $n$, the back propagation of loss is very time-consuming. Thus, an indispensable component is to sample hard negatives for both performance improvement and computation reduction. In the context of descriptor learning, the recent HarNet \cite{hard} method searches for the most difficult negative pair with reference to each achor positive pair. However, the hard negative sampling strategy in \cite{hard} is unable to fully explore the negative pairs because only negative pairs that share an element with the anchor positive pairs are considered. Due to the margin $m$, if the similarity between positive pairs and negative pairs is bigger than margin and then the derivation of triplet item will be 0. This will cause information loss, but triplet can help learn a better distribution of descriptors.

\subsection{The contrastive loss}

  The difference between the contrastive loss and the triplet loss is that triplet loss aims at comparing relative similarity between positive pairs and negative pairs, while contrastive loss only compares negative pairs with margin and pull positive pairs as close as possible. The general form of contrastive loss is formed as follows:
\begin{equation}
L_{contrastive}=\frac{1}{N}\sum_{i,j,k,l}([m+s(a_{k},n_{l})]_{+}-s(a_{i},p_{j}))
\end{equation}
In contrastive loss, the number of training data pairs grows quadratic with respect to training sample size. Therefore, it is much easier to sample the data pairs than triplet loss, and random sampling is often employed for contrastive loss-based learning. However, as shown in previous works, contrastive loss showed inferior performance to triplet loss in certain tasks. But \cite{DBLP:journals/corr/WuMSK17} argues that the inferior performance of contrastive learning is due to the inappropriateness of the random sampling strategy. When combined with the proposed negative sampling strategy, contrastive loss performs as well as triplet loss in the descriptor learning task. The common difficulty for triplet and contrastive loss is that margin cause a great impact in result. 

\begin{figure}[htb]
\begin{minipage}[b]{1.0\linewidth}
  \centering
  \centerline{\includegraphics[width=1.1\textwidth]{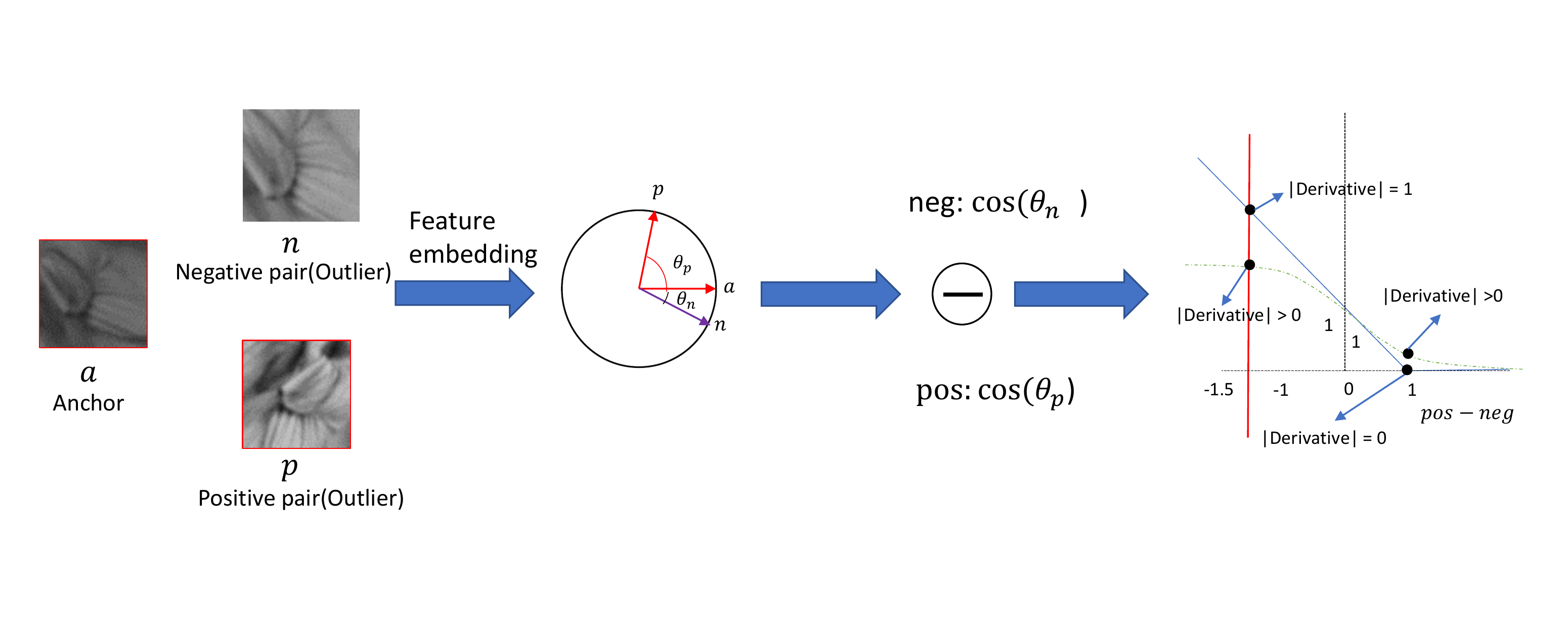}}
  \centerline{}\medskip
  \vspace{-1cm}
\end{minipage}
\vspace{-1cm}
\caption{Proposed descriptor learning model, the negative mining strategy is same as \cite{hard}. The features extracted from patches are located in embedding space and cosine similarity between positive pairs is $\cos\theta_{p}$ and negative pairs are $\cos\theta_{n_{1}}$ and $\cos\theta_{n_{2}}$.The curves in the right side represent triplet margin function(solid line),contrastive margin function(dash line) and our robust function(dash-doted line).}
\label{model}
\vspace{-0.2cm}
\end{figure}

As for triplet loss or contrastive loss, these methods all consist of positive pairs, negative pairs and a discriminative margin. The keypoint is the negative search strategy and margin choosing, the later one of which is tricky for metric learning. Thus we propose a robust loss without margin confusion which can keep more relative embedding information as well as applying cosine similarity for our metric learning which is similar to cosine face \cite{cosface},cause the angle distace is closer to original embedding distribution as a hypersohere. As shown in Fig.\ref{derivative}, assuming margin values for  triplet loss and contrastive loss are 1 and 0 separately which are common margin choice. Due to the margin strategy, there is always a part of selected triplet or contrastive items with no derivation, also the selected negative pairs are too sparse, resulting in a large selection bias in the sampling process. Explained in \cite{Yu2010RelaxedCA}, robust regression and classification problem often requires non-convex loss function which can prevent scalable and global training where a natural approach to implement it is cutting out loss vales that exceed threshold, which is similar to triplet and contrastive loss. However, a relaxation of this kind of 'clipping loss' can improve robustness. As for our embedding metric learning problem, we can adopt this intuition. 

Thus, we propose a robust version of triplet loss which can offset bias problem to some extent. We consider that selected positive pairs $(a_{i},p_{i}$ and negative pairs $(a_{i},n_{i})$ should be more important when when their similarity are high and vice versa given less weight rather than set their derivation to be 0. However, we observed that a considerable amount false negative labels exiting which should be positive pairs but marked as negative which is shown in Fig.\ref{model}, therefore, we put less weight to triplet items when  the cosine distance between negative pairs $(a_{i},n_{i})$ is much more bigger than positive pairs $(a_{i},p_{i}$, the derivation of which should be similar to symmetrical arch bridges as shown in Fig.\ref{derivative} (d). Our method performs obviously better when training data by small batch where the effect of bias influences more and reach the best for bigger batch size, which will be discussed later.

We apply the same positive pairs and negative pairs sampling strategy as \cite{hard} and the features generated by networks are 128-D and euclidean normalized with length 1. We define a search procedure $S_{i}$ starting with anchors $(a_{i}$ and we search for all of the  positive and negative items related to $a_{i}$. With regard to a batch which consists of $N$ pairs of matched patches $a$ and $p$ from different given 3D point of view of descriptors,the size of descriptors matrices A and P are $N \times 128$, $N\times N$ cosine cosine similarity matrix $\bm{D}=A \times P^T$, so there is only one positive item $pos_{i}=d(a_{i},p_{i})$ in the diagonal of $\bm{D}$ for each search $S_{i}$. The goal is to find the closest negative item with respect to $a_{i}$ and $p_{i}$.As we know, the bigger the gap between two features, the smaller the cosine similarity is. Please refer to \cite{hard} for more sampling details, the formula is organized as follow:

\begin{align}
\begin{split}
&\qquad \qquad \quad \bm{D}(i,j)=\cos \theta_{i,j}=a_{i} \cdot p_{j},\\
&\qquad \qquad \quad pos_{i} = \bm{D}(i,i),\\
&\qquad \qquad \quad neg_{i} = \bm{D}(r_{i},c_{i}),\\
&\qquad \qquad \quad (r_{i},c_{i})={\rm argmax}_{k,l}\bm{D}(k,l),\\
&\qquad \quad \qquad s.t.\quad k,l=1...n,\\
&\qquad \quad \qquad \quad \quad \  \{k = i \vee l = i\}= True,\\
&\qquad \quad \qquad \quad \quad \  k\neq l.\\
\end{split}&
\end{align}
Finally, the triplet items are fed into loss function formed as follow,and our goal is to minimize this loss for each batch. Also, the derivation of $L_{i}$ with respect to $(pos_{i}-neg_{i})$ is even function as explained above in Fig.\ref{derivative}, and as for better description,Fig.\ref{derivative} demonstrate the absolute value of derivation.
\begin{align}
\begin{split}
&\qquad \qquad L_{robust}=\frac{1}{N}\sum_{i=1}^{N}{(1 - \tanh(pos_{i}-neg_{i}))},\\
&\qquad \qquad L_{i} = 1 - \tanh(pos_{i}-neg_{i}),\\
&\qquad \qquad \frac{\partial L_{i}}{\partial (pos_{i}-neg_{i})} = \tanh^2 ((pos_{i}-neg_{i})) -1 .
\end{split}&
\end{align}

\subsection{Network structure}

Following \cite{hard}, we adopt the L2Net \cite{8100132} architecture as our main network. The network consists of two parts, the main feature extraction network and the linear decision layer which reduces the feature dimension. For fair comparison, we also make slight modifications by adding 0.3 dropout layer above the bottom layer of network as \cite{hard}. For an $32 \times 32$ normalized single-channel patch, the output is a L2 normalized 128-dimensional feature vector. It is worth noting that the whole feature extraction network is built by full convolutional layers, downsampling by two-stride convolutional net. Also, and there is a BN layer and a Relu activation layer in every layer except the last layer, except the bottom layer with no Relu layer. And the whole network is trained with the proposed negative sampling procedure and corresponding loss functions.

\section{EXPERIMENTAL RESULTS}
We train and test our RAL-Net descriptor on the  Brown dataset and test the patch verification performance on Brown dataset. In addition, we use the models trained on the Brown dataset to test its generalization abilities in patch verification, patch matching, and patch retreival on the Hpatches dataset. Finally, we apply our RAL-Net descriptor on the Wide Baseline Stereo dataset to test its invariance properties.

\subsection{Brown dataset}
\begin{figure}[htb]
\begin{minipage}[b]{0.31\linewidth}
  \centerline{\includegraphics[width=3cm]{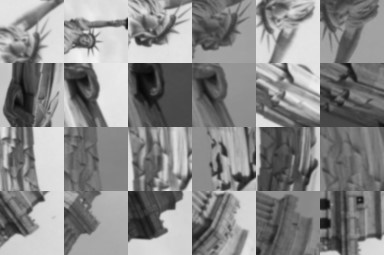}}
  \centerline{(a) Liberty}\medskip
\end{minipage}
\hfill
\begin{minipage}[b]{0.31\linewidth}
  \centerline{\includegraphics[width=3cm]{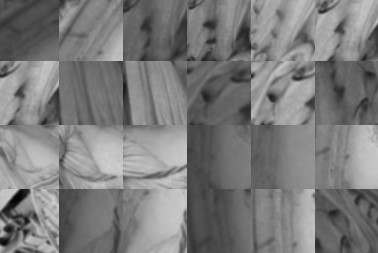}}
  \centerline{(b) Notredame}\medskip
\end{minipage}
\hfill
\begin{minipage}[b]{0.31\linewidth}
  \centerline{\includegraphics[width=3cm]{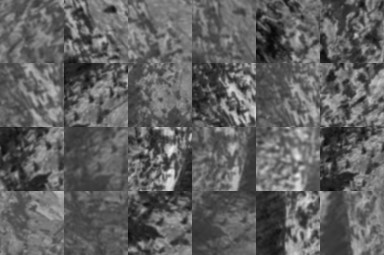}}
  \centerline{(b) Yosemite}\medskip
\end{minipage}
\caption{Subsets of Brown}
\label{Brown}
\end{figure}
The Brown dataset is the most popular local descriptor learning dataset, which contains three subsets of images taken from different places, including Liberty, Notredame and Yosemite. Keypoints are firstly detected by Difference of Gaussians (DOG) \cite{Lowe2004} and then vefified with ground truth 3-D view. The patches are extracted around the keypoint locations and are normalized by scale and orientation calculated during keypoint detection. There are about 400k classes of patch pairs with 64$\times$64 size, extracted from different sequences. In practice, the size of 64$\times$64 is unnecessary, and we resized the patches to 32$\times$32 by linear cubic interpolation. 

\paragraph{Training setting}

On the three Brown subsets, we trained our RAL-Net descriptor in different setting with different training sample size and batch size. In the first setting, we only extracted 200K pairs in total for each subsect repectively. In this setting, we compared the performance of our descriptor with the state-of-the-art HardNet trained with a small batchsize 128. The performance with a small batch size can desmonstrate the effectiveness of the negative sampling strategies. We applied the training strategy different from \cite{hard} and \cite{8100132}, which trains data for 50 epochs with learning rate linearly decreasing to 0 in the end. We choose Stochastic Gradient Descent (SGD) as our optimizer and we set the initial learning rate to be 10, and the rest momentum to be 0.9 and weight decay to be 0.0001. In addition, we have tried Adam optimizer and it converge faster than SGD, however SGD can achieve a better result with well chosen training parameters. 

In the second setting, which is the standard setting, we extracted 5000K pairs and trained descriptor with batch size 512. As for HardNet, we trained it using batch sizes 512 and 1024, as the performance of HardNet is more sensitive to batch size. In this setting, due to big amount of data, the training is done within 10 epoch which is much less than strategy one but the other training aspects are the same as strategy one. In order to compare our RAL-Net descriptors, we also apply the same training strategy on HardNet and cite several result of recent works. Following previous works, we also applied data augmentation by random filpping and $90^\circ$ rotation in both training settings. 

\label{ssec:subhead}
\begin{figure}[htb]
\begin{minipage}[b]{0.48\linewidth}
  \centerline{\includegraphics[width=6cm]{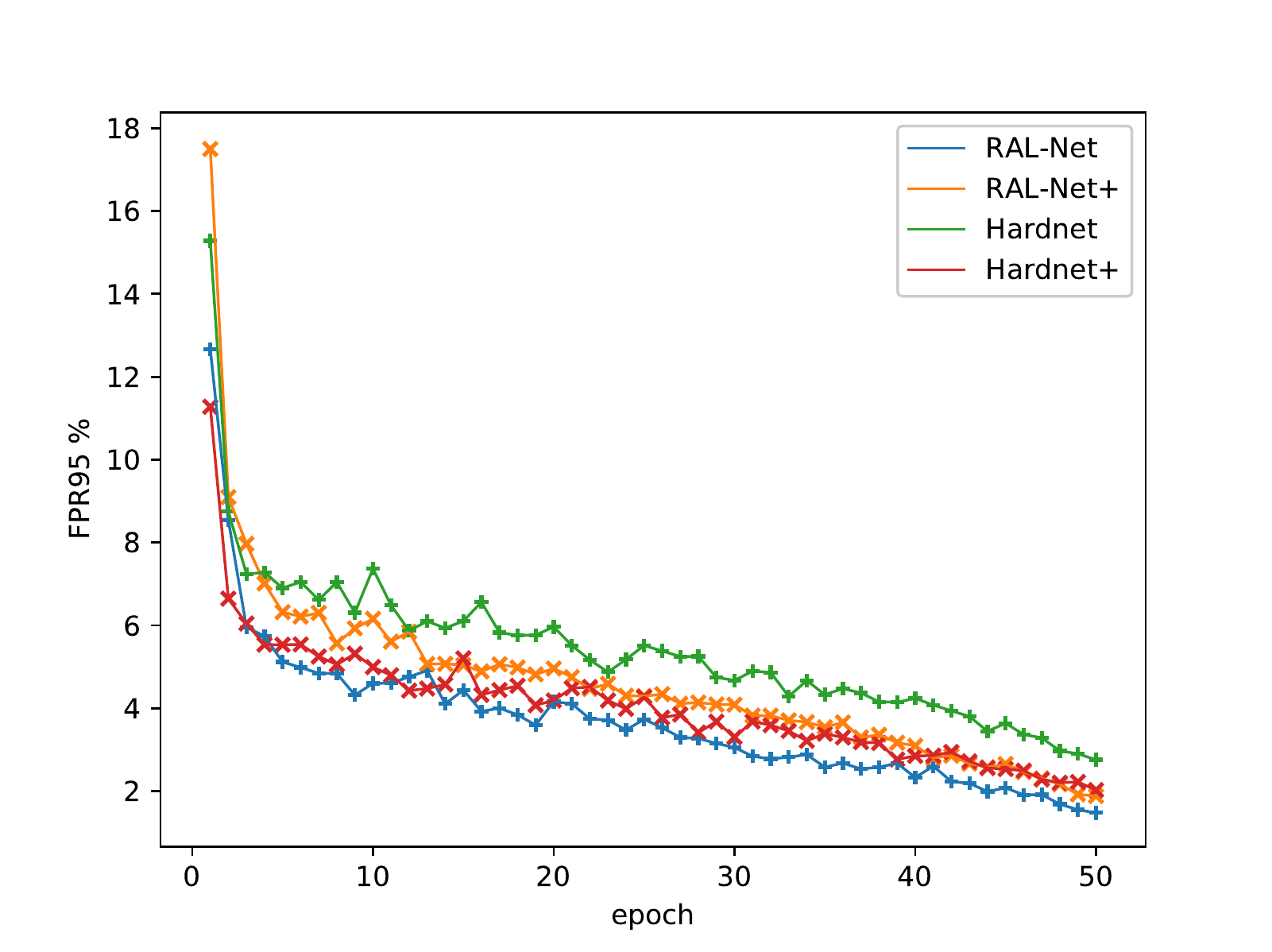}}
  \centerline{(a) Training strategy 1}\medskip
\end{minipage}
\hfill
\begin{minipage}[b]{0.48\linewidth}
  \centerline{\includegraphics[width=6cm]{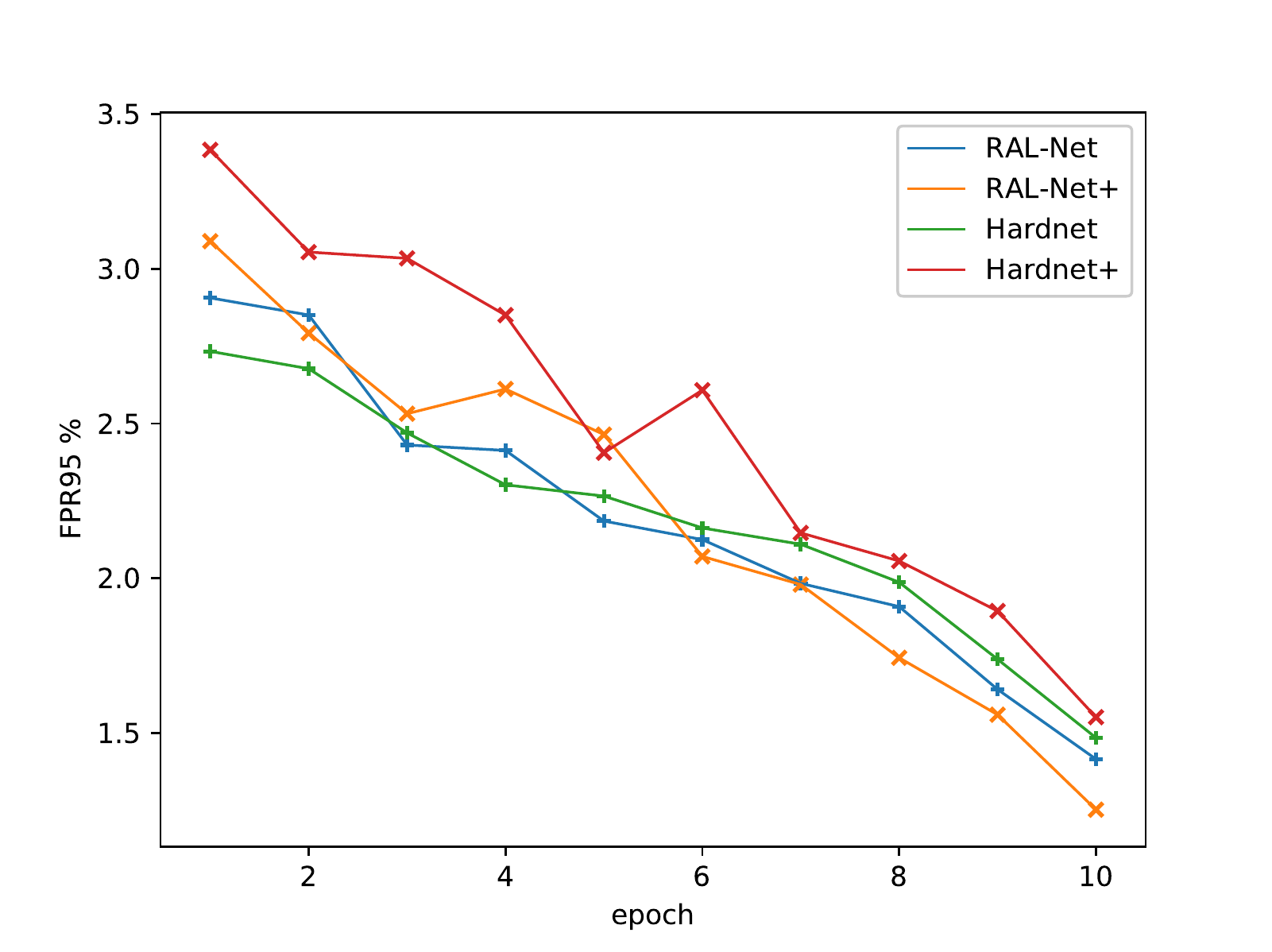}}
  \centerline{(b) Training strategy 2}\medskip
\end{minipage}
\caption{The curves describe the result of FPR95 of Hardnet and our RAL-Net tested in Brown dataset.X axis is training epochs and Y is the percent value of FPR95.}
\label{loss}
\vspace{-0.3cm}
\end{figure}

\begin{table}[h]
\large
\centering
\renewcommand\arraystretch{1.4}
 \caption{Descriptor performance on Brown dataset for patch verification. False positive rate at 95\% true positive rate is displayed. Results of the best are in bold and "+" suffix represent training implemented by data augmentation of random flipping and $90^\circ$ rotating. }
 \resizebox{1\textwidth}{!}{
 \begin{tabular}{ccccc}
  \toprule
 Training{ } { } & \underline{Notredame{ } { } Yosemite{ } { }} & \underline{Liberty { } { } Yosemite{ } { }} & \underline{Liberty { } { } Notredame{ } { }} & { } { }Mean  \\
 Test & Liberty & Notredame & Yosemite & FPR \\
 \midrule
   SIFT\cite{Lowe2004} & 29.84 & 22.53 & 27.29 & 26.55  \\
   MatchNet \cite{7298948} & 7.04 { } { } 11.47 & 3.82 { } { } 5.65 & 11.6 { } { } 8.7 & 8.05  \\
   L2Net\cite{8100132} & 3.64 { } { } 5.29 & 1.15 { } { } 1.62 & 4.43 { } { } 3.30 & 3.23  \\
 L2Net${+}$\cite{8100132} & 2.36 { } { } 4.70 & 0.72 { } { } 1.29 & 2.57 { } { } 1.71 & 2.22\\
 CS L2Net\cite{8100132} & 2.55 { } { } 4.24 & 0.87 { } { } 1.39 & 3.81 { } { } 2.84 & 2.61\\
 CS L2Net${+}$\cite{8100132} & 1.71 { } { } 3.87 & 0.56 { } { } 1.09 & 2.07 { } { } 1.3 & 1.76\\
  HardNetNIPS \cite{hard} { } { } & 3.06 { } { } 4.27 & 0.96 { } { } 1.4 & 3.04 { } { } 2.53 & 2.54\\
 HardNet+NIPS \cite{hard} { } { } & 2.28 { } { } 3.25 & 0.57 { } { } 0.96 & 2.13 { } { } 2.22 & 1.9 \\
 \midrule
 \multicolumn{5}{c}{Traing strategy 1: 200K training pairs for each subset, batch size 128} \\
 \midrule
 HardNet$_{128}${ } { } &2.07  { } { } 3.70  &0.77  { } { }1.22  & 3.79 { } { } 3.33 &2.48  \\
 HardNet$_{128}$+{ } { } & 2.46 { } { } 3.55  & 0.73 { } { } 1.67 & 3.54 { } { } 3.40 & 2.56 \\
RAL-Net$_{128}$(ours){ } { } & \textbf {1.46} { } { } \textbf{2.63} &\textbf {0.51} { } { } \textbf{0.91} & \textbf{1.95} { } { }\textbf {1.40} &\textbf {1.48} \\
 RAL-Net$_{128}$+(ours){ } { } &  {1.81} { } { }{ 3.80} & {0.55} { } { } {1.01} & {1.96} { } { } {2.18} & {1.89} \\ 
 \midrule
 \multicolumn{5}{c}{Traing strategy 2: 5000k training pairs for each subset, batch size 512} \\
 \midrule
 HardNet$_{512}$  { } { } & 1.54 { } { } 2.56 & 0.63 { } { } 0.92 & 2.65 { } { } 2.05 & 1.73\\
 HardNet$_{512}$+  { } { } & 2.53 { } { } 2.69 & 0.54 { } { } 0.83 & 2.49 { } { } 1.70 & 1.80 \\
  HardNet$_{1024}$  { } { } & 1.47 { } { } 2.67 & 0.62 { } { } 0.88 & 2.14 { } { } 1.65 & 1.57\\
 HardNet$_{1024}$+  { } { } & 1.49 { } { } 2.51 & 0.53 { } { } 0.78 & 1.96 { } { } 1.84 & 1.51 \\
 RAL-Net$_{512}$(ours) { } { } & {1.44} { } { } {2.60} & {0.48} { } { } {0.77} & {1.77} { } { } {1.43} & {1.42}\\
 RAL-Net$_{512}$+(ours) { } { } &\textbf {1.30} { } { }\textbf {2.39} &\textbf {0.37} { } { }\textbf {0.67} &\textbf {1.52} { } { }\textbf {1.31} &\textbf {1.26}\\
  \bottomrule
 \end{tabular}
 }
 \label{Brown result}
\vspace{-0.2cm}
\end{table}

\paragraph{Overall evaluation}

The descriptors are trained on one subset and tested on the rest two subsets. As for evaluation, tested subset contains 100k pairs of patches for each subset with 50K matched and 50K unmatched labels. We follow the evaluation protocol \cite{5432199} and give the results of false positive rate FPR at the recall of $95\%$ true positive rate TPR (FPR95).The training precision alone training epoch is demonstrated in Fig.\ref{loss} and the results are shown in Table.\ref{Brown result}, the best results are shown in bold.

Obviously, our RAL-Net generate the overall best results among all of the representative descriptors as well as the best among the testing subsets. Deep model-based descriptors have surpassed far more than hand-crafted ones, and focus on comparison between our descriptor and the HardNet descriptor. 

In the first training setting, it is interesting that RAL-Net descriptor achieves better results than the HardNet descriptor when both of them are trained with 200K pairs with batch size of 128. Furthermore, our descriptor trained with less data achieves comparable results as Hardnet trained on 5000K data pairs with 512 batch size. We can also notice that data augmentation shows no better effect for small training sample size and even slightly worsen the performance due to the increasing difficulties of negative sampling and less training data with bigger bias. However, the results on the small training dataset and small batchsize verifies the effectiveness of our proposed robust loss training.

In the second setting, we obtain the best results on 5000K data pairs with a batch size of 512. Even in the training without data augmentation, RAL-Net exceeds Hardnet with batch size 512 as well as 1024. It is worth noting that Hardnet descriptor gets improved when enlarging batch size from 512 to 1024, while increasing batchsize from 512 to 1024 leads to almost no enhancement for our descriptor. Thus, we only report the results of our Ral-Net trained with batch size of 512. The experimental results confirm the efficiency and validity of our RAL-Net descriptor. For the rest of experiments, we test HardNet and our descriptor on other datasets by training them on 5000K data pairs with 512 batch size on the Liberty subset.

\paragraph{Ablation studies}

Proving the effectiveness of angular and robust form separately with respect to our RAL loss. We set four variants based on our training strategy 1 without augmentation due to limited time, which are HardNet with angular embedding, HardNet with robust form, original HardNet and our RAL-Net respectively. The result is shown in Table.\ref{abalation}. The results indicate that both the angular distance and the robust loss function contribute to the overall performance and the combination of them achieves state-of-the-art performance. 
\begin{table}[h]
\large
\centering
\renewcommand\arraystretch{1.4}
 \caption{Comparision between different combinations. }
 \resizebox{1\textwidth}{!}{
 \begin{tabular}{ccccc}
  \toprule
 Training{ } { } & \underline{Notredame{ } { } Yosemite{ } { }} & \underline{Liberty { } { } Yosemite{ } { }} & \underline{Liberty { } { } Notredame{ } { }} & { } { }Mean  \\
 Test & Liberty & Notredame & Yosemite & FPR \\
 \midrule
 \multicolumn{5}{c}{Traing strategy 1: 200K training pairs for each subset} \\
 \midrule
   RAL-Net{ } { } & \textbf {1.46} { } { } \textbf{2.63} &\textbf {0.51} { } { } \textbf{0.91} & \textbf{1.95} { } { }\textbf {1.40} &\textbf {1.48} \\
  HardNet/ angular embedding { } { } & 1.63 { } { } 3.26  & 0.56 { } { } 1.24 & 2.87 { } { } 2.02 & 1.93 \\
  HardNet / robust form{}{}&1.67  { } { } 3.27  &0.57  { } { }1.06  & 2.51 { } { } 2.15 & 1.87\\
  HardNet {}{}&2.07  { } { } 3.70  &0.77  { } { }1.22  & 3.79 { } { } 3.33 & 2.48\\
  L2Net/contrastive {}{}&3.52  { } { } 7.83  &1.63  { } { }2.75  & 7.36 { } { } 6.68 & 4.96\\
  \bottomrule
 \end{tabular}
 }
 \label{abalation}
\end{table}

\subsection{Descriptor generalization ability on Hpatches dataset}
\begin{figure}[htb]
\begin{minipage}[b]{1\linewidth}
  \centering
  \centerline{\includegraphics[width=1\textwidth]{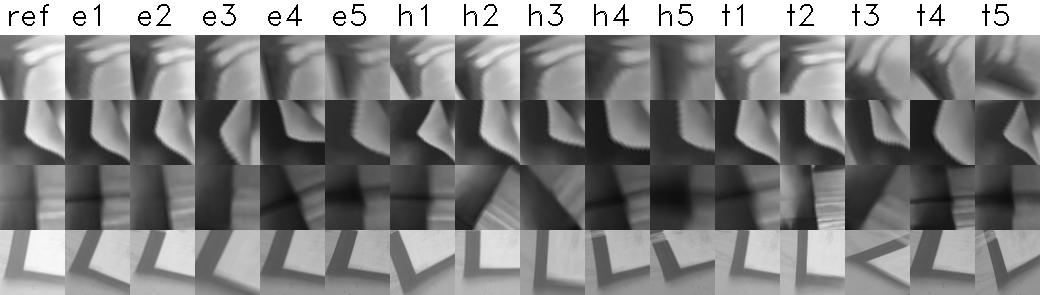}}
  \centerline{}\medskip
\end{minipage}
\vspace{-0.5cm}
\caption{Hpatches patches image. For each reference patch, there are 5 random geometric changing patches for three different changing range, which can be classified to e(easy), h(hard) and t(tough).}
\label{Hpatch data}
\end{figure}

Recently, Hpatches, a new local descriptor evaluation benchmark, provides a huge dataset and an evaluation criterion for modern descriptors. This dataset consists of $65 \times 65$ pixel size of patches extracted from 116 sequences which originate from 6 images. Different from the widely used Brown dataset, Hpatches contains more diversity and noisy changes. The keypoints of this dataset are detected by DOG, Hessian, and Harris detectors from reference images which are then applied to reproject the three different geometric noisy image sequences of easy, hard and tough. A small fraction of the dataset is shown in Fig.\ref{Hpatch data}.

\begin{figure}[htb]

\begin{minipage}[b]{0.5\linewidth}
  \centerline{\includegraphics[width=6.2cm]{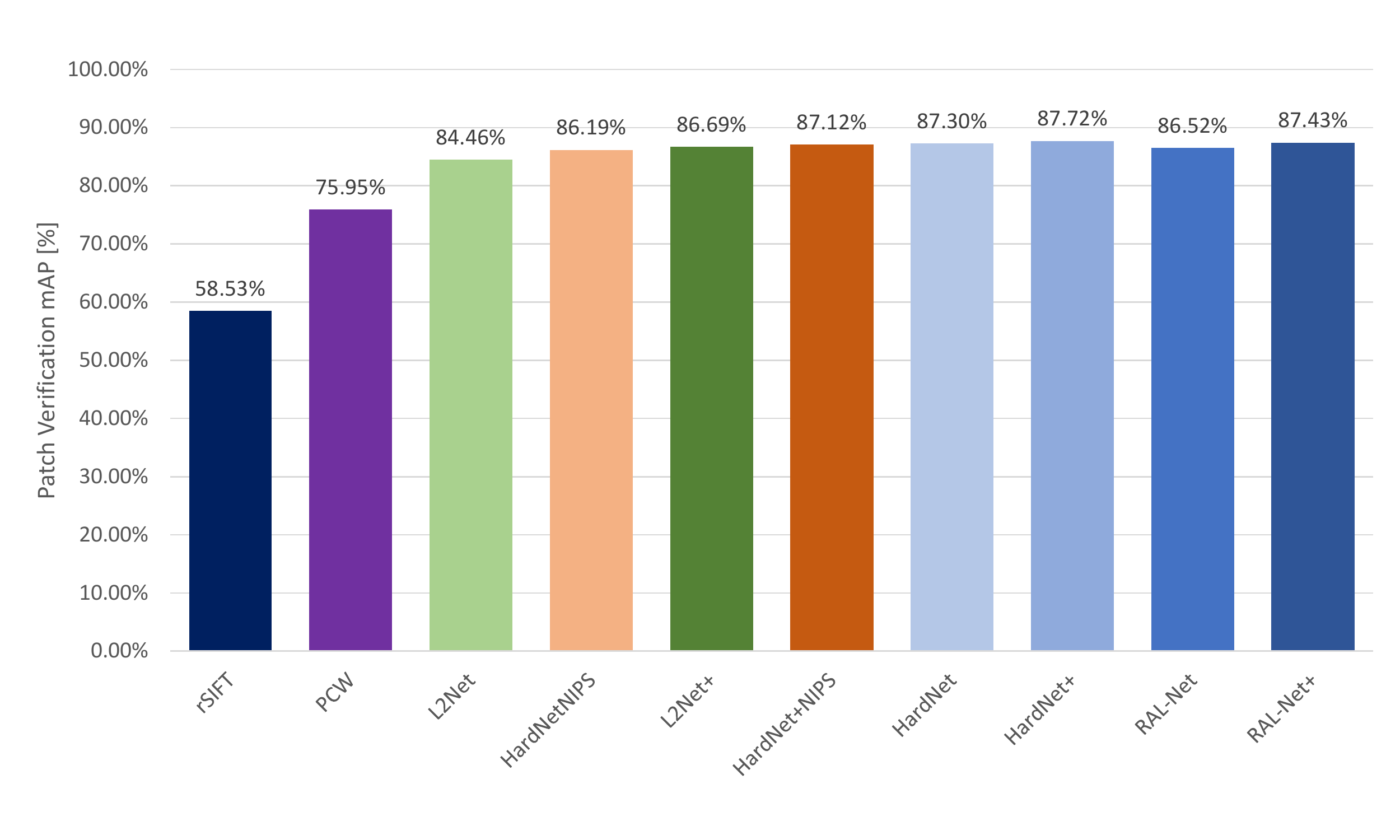}}
  \centerline{(a) Patch Verification}\medskip
\end{minipage}
\hfill
\begin{minipage}[b]{0.5\linewidth}
  \centerline{\includegraphics[width=6.2cm]{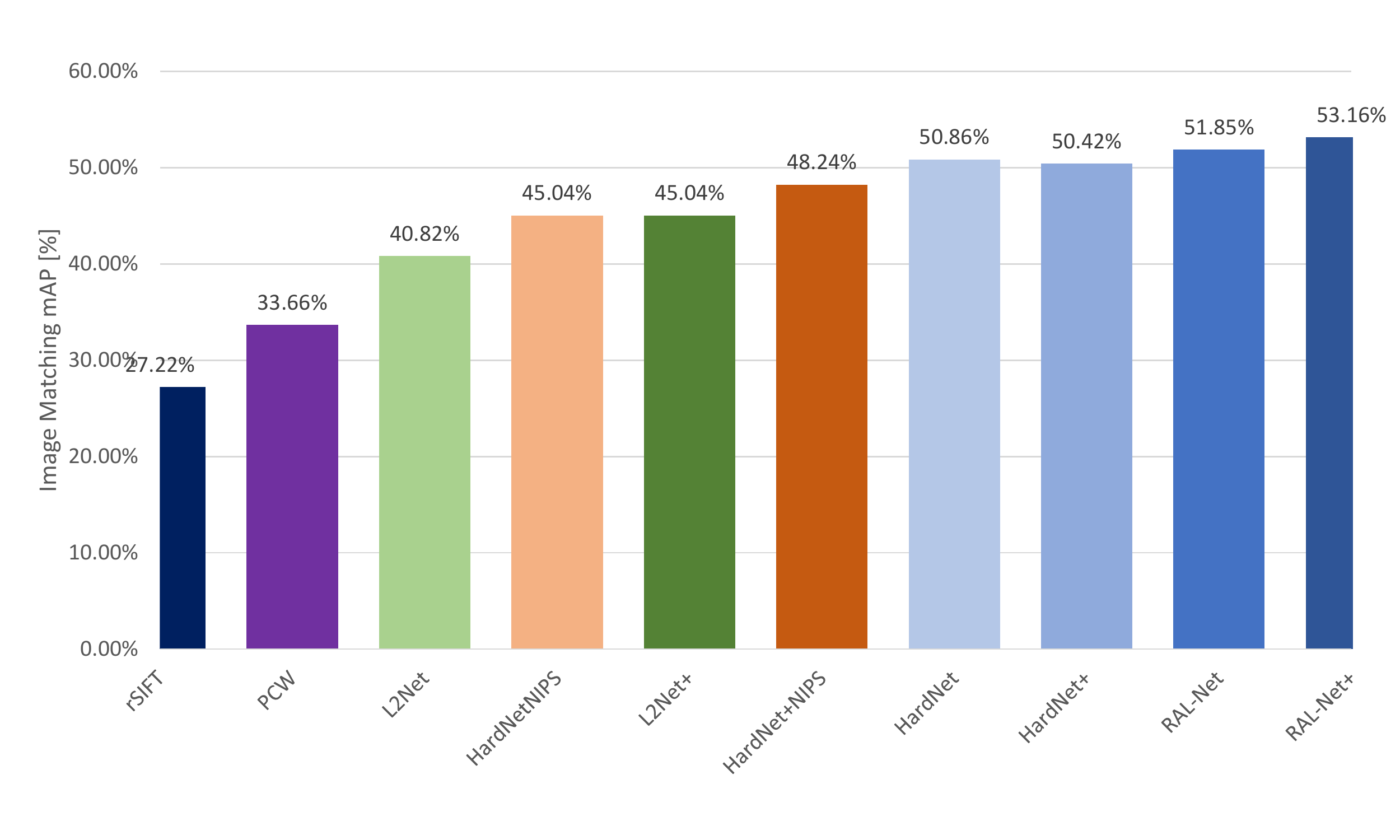}}
  \centerline{(b) Patch Matching}\medskip
\end{minipage}

\begin{minipage}[b]{1.0\linewidth}
  \centering
  \centerline{\includegraphics[width=6.2cm]{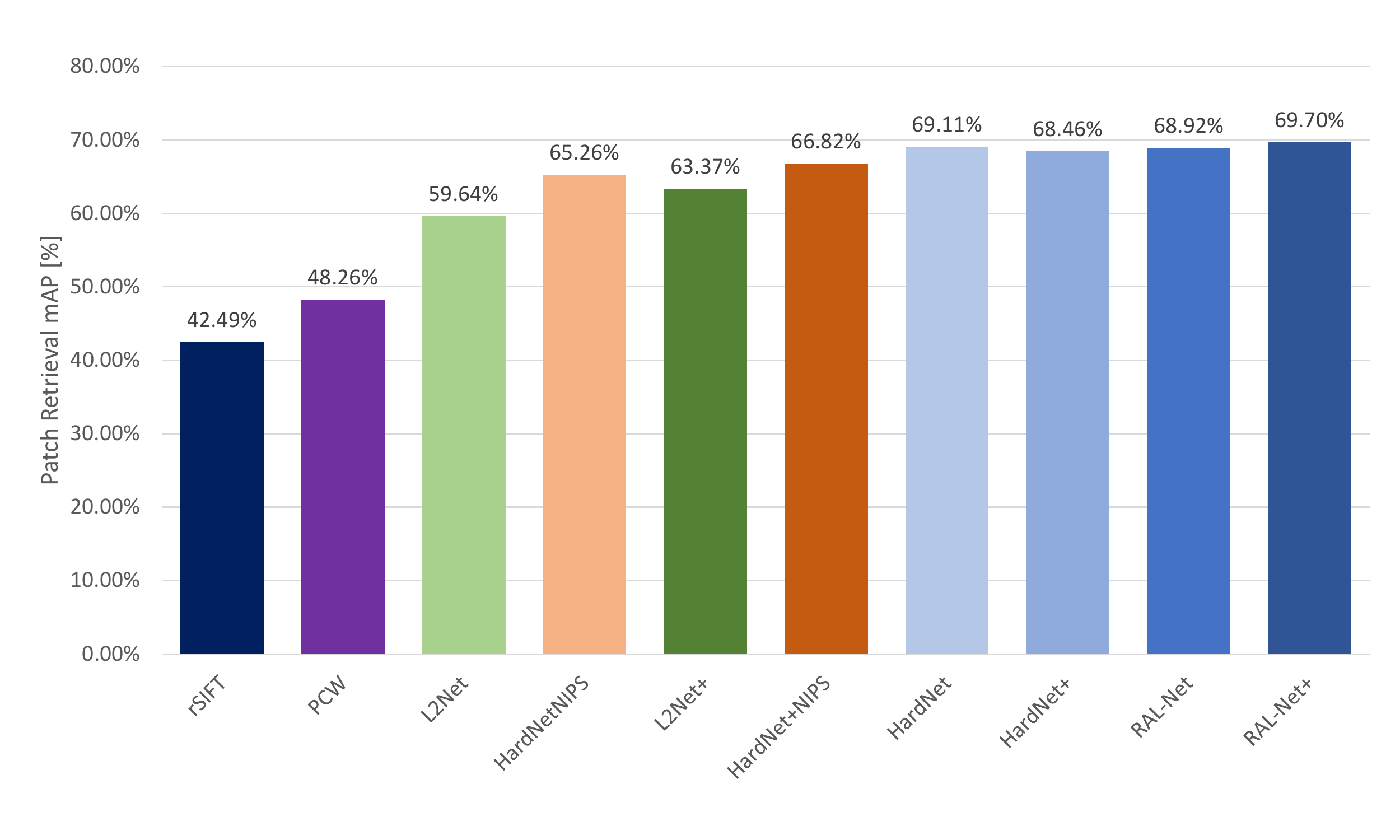}}
  \centerline{(c) Patch Retrieval}\medskip
\end{minipage}
\vspace{-0.5cm}
\caption{Descriptors performance on three tasks}
\label{Hpatch result}
\end{figure}

To comprehensively test the generalization abilities of descriptors, Hpatches also propose three different tasks, including patch verification, image matching, and patch retrieval. First, patch verification is used to verify whether two patches match or not by confidence scores. As for a patches pair set $P= \{(x_{i},x_{i}^{\prime}),y_{i}),i =1,...,N)\}$ , consisting of positive pairs and negative pairs $(x_{i},x_{i}^{\prime})$ with labels $(y_{i}=1,-1)$, we calculate the average precision by the ranked confidence scores. The mean average precision (mAP) for all of the rank is finally used as the evaluation criterion. Second, image matching is similar to patch verification, in which we are given patches collection $L_{k}=(x_{k,i},i = 1,...,N)$, where $L_{r}$ is from the reference image and $L_{t}$ is from the target image. With respect to $x_{r,i}$ from $L_{r}$, we aim to find the maximum matching $x_{t,j}$ from $L_{t}$ and get the related index $\{\sigma _{i}, i = 1,...,N\}$ of finding $x_{t,j}$. After finding all of the matching, we consider if the found $x_{t,j}$ corresponds to $x_{r,i}$ with a ground truth label, and we get the mathching set $M=\{y_{i}= 2[\sigma _{i}\overset{?}{=}i]-1\}$ by whether the found patch is matched with the label. Similar to the first task, we calculate the mAP for AP of set M for all ranks. The final task is patch retrieval, which considers these retrieved patches from the the matched images of reference images with a large proportion of distraction, and returns AP of the collection of labels ranked by confidence scores. For more protocol details, please refer to \cite{DBLP:journals/corr/BalntasLVM17}. The result is shown in Fig.\ref{Hpatch result}.

Hpatches evaluation protocol considers many different aspects of patches from different view points and different illumination as well as patches of intra-class and inter-class, which are implemented from three degree of difficult sequences separately. In order to make a clearer demonstration, we just give the average performance of all different factors. Actually, these descriptors which obtain better average result shown in Fig.\ref{Hpatch result} also perform better on these different child factors respectively. In terms of the result, RAL-Net generates the best results on the image matching task and the patch retrieval task, and a little behind HardNet in patch verification task. We can also observe that for the hardest image matching task, our descriptor with data augmentation shows an obvious improvement over the competitors. Overall, our descriptor gives almost the same  results as L2Net and HardNet. This might be due to the different distributions between the Hpatches data and the Brown data, which needs further consideration and new learning algorithms such as transfer learning.

\subsection{WxBS testing}
\begin{figure}[htb]

\begin{minipage}[b]{1\linewidth}
  \textbf{\qquad \quad G \qquad \qquad \quad A \qquad \qquad \quad L \qquad \qquad \quad Map2ph \qquad \qquad S}\par\medskip
  \centerline{\includegraphics[width=12cm]{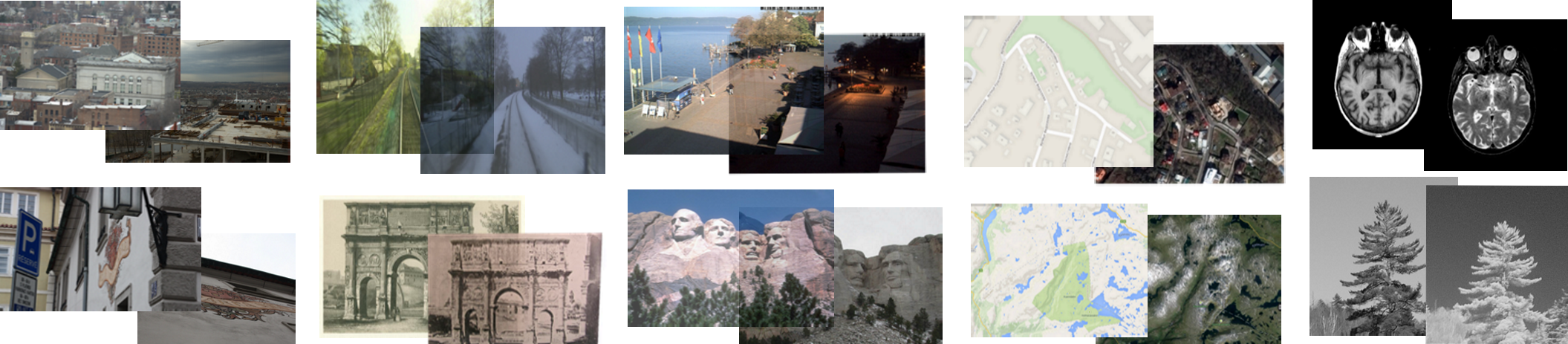}}
  \centerline{}\medskip
\end{minipage}
\vspace{-1cm}
\caption{Examples of WxBS dataset}
\label{Wxbs data}
\end{figure}

\begin{figure}[htb]

\begin{minipage}[b]{1\linewidth}
  \centerline{\includegraphics[width=12cm]{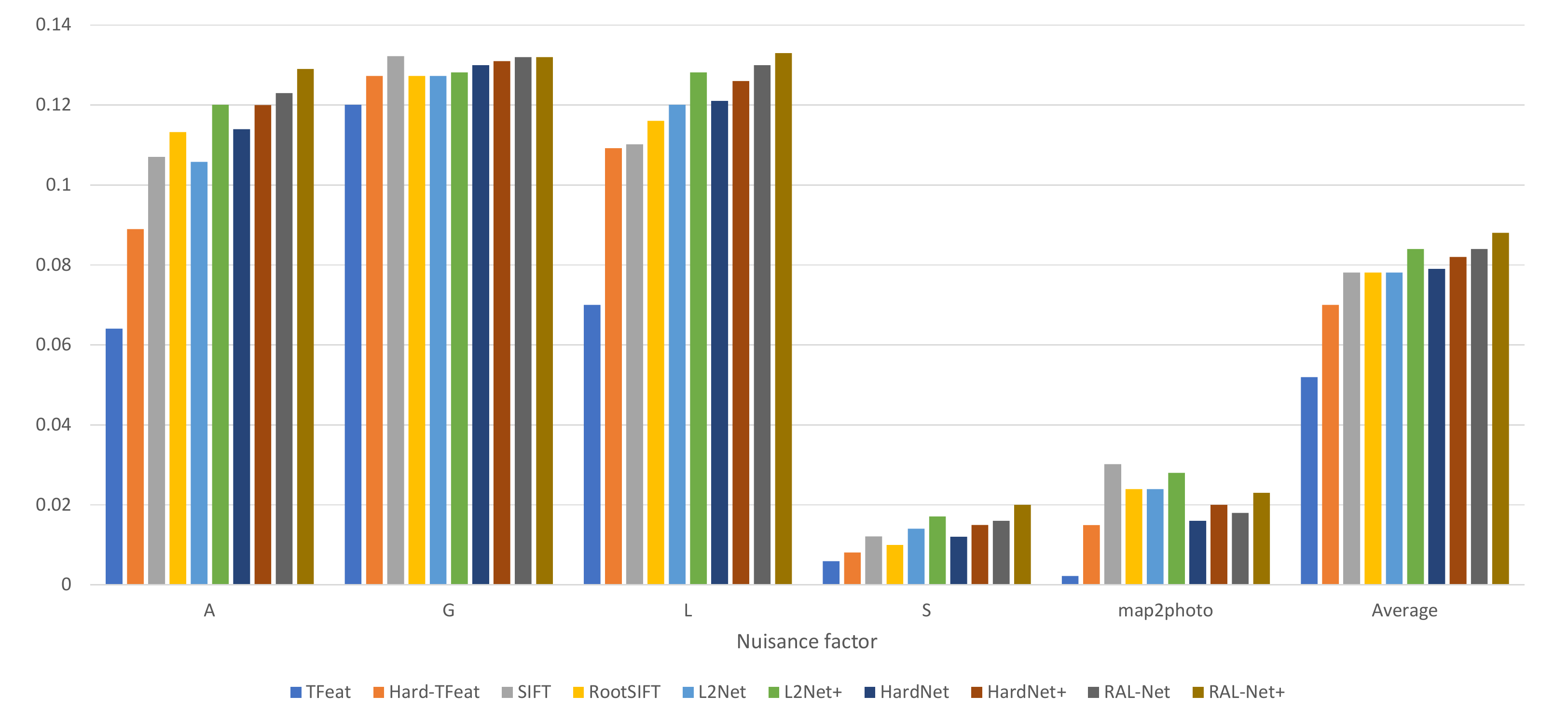}}
  \centerline{}\medskip
\end{minipage}
\vspace{-1cm}
\caption{Descriptors performance on three tasks}
\label{Wxbs result}
\vspace{-0.3cm}
\end{figure}

In order to test the performance of our RAL-Net descriptor in a hard environment with various changing factors, we apply our descriptor on Wide Baseline Stereo \cite{DBLP:journals/corr/MishkinMPL15}. The dataset consists of three different tasks, which are Appearance (A) by environment change, Geometry (G) with different view point, scale variance, etc., Illumination (L) influenced by brightness or image intensity, and Sensor (S) consisting of different type of data. With local feature detected by maximally stable extremal regions (MSER), Hessian-Affine and FOCI, each local patch is matched perfectly with the reference image. In addition, the evaluation metric is the same as the image matching task as Hpatches. The image example is shown as Fig.\ref{Wxbs data} and the result is shown in Fig.\ref{Wxbs result}.

RAL-Net performs the best on average and shows a distinguished performance on Appearance and Illumination distraction matching task. For the Geometry task, all of the task performs almost at the same level. It is also worth noticing that SIFT and RootSIFT do not fall behind these learning-based descriptors and even perform better on the task Geometry due to their scale-invariant characteristics. But,as we can observe in Fig.\ref{Wxbs data}, all of the descriptor reach a quite bad performance in S and Map2ph task due to this kind of data not exiting in Brown dataset.

\section{Conclusions}
In this paper, we suggest a robust angular training loss named RAL-Net for deep embedding learning without sensitive parameters to further improve the performance of local descriptor learning,where the similarity between descriptors is defined as cosine distance, it is based on the idea of smooth the margin of triplet by giving different importance to triplet items with regard to the difference between the similarity of positive pairs and the similarity between chosen negative pairs. The loss can learn more information from limited data and performs better if with larger training data and relax the effect that false labels exits in training dataset. We test DigNet on typical Brown dataset, Hpathches dataset and W1BS dataset for diverse tasks verification and our RAL-Net have shown a superiority over existing local descriptors.

\bibliographystyle{splncs}
\bibliography{0063.bbl}

\begin{thebibliography}{10}

\bibitem{8100132}
Tian, Y., Fan, B., Wu, F.:
\newblock L2-net: Deep learning of discriminative patch descriptor in euclidean
  space.
\newblock In: 2017 IEEE Conference on Computer Vision and Pattern Recognition
  (CVPR). (2017)  6128--6136

\bibitem{hard}
Mishchuk, A., Mishkin, D., Radenovic, F., Matas, J.:
\newblock Working hard to know your neighbor’s margins: Local descriptor
  learning loss.
\newblock In Guyon, I., Luxburg, U.V., Bengio, S., Wallach, H., Fergus, R.,
  Vishwanathan, S., Garnett, R., eds.: Advances in Neural Information
  Processing Systems 30.
\newblock Curran Associates, Inc. (2017)  4829--4840

\bibitem{NIPS2016_6487}
Choy, C.B., Gwak, J., Savarese, S., Chandraker, M.:
\newblock Universal correspondence network.
\newblock In Lee, D.D., Sugiyama, M., Luxburg, U.V., Guyon, I., Garnett, R.,
  eds.: Advances in Neural Information Processing Systems 29.
\newblock Curran Associates, Inc. (2016)  2414--2422

\bibitem{4270197}
Philbin, J., Chum, O., Isard, M., Sivic, J., Zisserman, A.:
\newblock Object retrieval with large vocabularies and fast spatial matching.
\newblock In: 2007 IEEE Conference on Computer Vision and Pattern Recognition.
  (2007)  1--8

\bibitem{790410}
Lowe, D.G.:
\newblock Object recognition from local scale-invariant features.
\newblock In: Proceedings of the Seventh IEEE International Conference on
  Computer Vision. Volume~2. (1999)  1150--1157 vol.2

\bibitem{Lowe2004}
Lowe, D.G.:
\newblock Distinctive image features from scale-invariant keypoints.
\newblock International Journal of Computer Vision \textbf{60} (2004)  91--110

\bibitem{2014arXiv1405.5769F}
{Fischer}, P., {Dosovitskiy}, A., {Brox}, T.:
\newblock {Descriptor Matching with Convolutional Neural Networks: a Comparison
  to SIFT}.
\newblock ArXiv e-prints (2014)

\bibitem{7410379}
Simo-Serra, E., Trulls, E., Ferraz, L., Kokkinos, I., Fua, P., Moreno-Noguer,
  F.:
\newblock Discriminative learning of deep convolutional feature point
  descriptors.
\newblock In: 2015 IEEE International Conference on Computer Vision (ICCV).
  (2015)  118--126

\bibitem{7298948}
Han, X., Leung, T., Jia, Y., Sukthankar, R., Berg, A.C.:
\newblock Matchnet: Unifying feature and metric learning for patch-based
  matching.
\newblock In: 2015 IEEE Conference on Computer Vision and Pattern Recognition
  (CVPR). (2015)  3279--3286

\bibitem{2015arXiv150403641Z}
{Zagoruyko}, S., {Komodakis}, N.:
\newblock {Learning to Compare Image Patches via Convolutional Neural
  Networks}.
\newblock ArXiv e-prints (2015)

\bibitem{6718113}
Simonyan, K., Vedaldi, A., Zisserman, A.:
\newblock Learning local feature descriptors using convex optimisation.
\newblock IEEE Transactions on Pattern Analysis and Machine Intelligence
  \textbf{36} (2014)  1573--1585

\bibitem{Bentley:1975:MBS:361002.361007}
Bentley, J.L.:
\newblock Multidimensional binary search trees used for associative searching.
\newblock Commun. ACM \textbf{18} (1975)  509--517

\bibitem{7298682}
Schroff, F., Kalenichenko, D., Philbin, J.:
\newblock Facenet: A unified embedding for face recognition and clustering.
\newblock In: 2015 IEEE Conference on Computer Vision and Pattern Recognition
  (CVPR). (2015)  815--823

\bibitem{5432199}
Brown, M., Hua, G., Winder, S.:
\newblock Discriminative learning of local image descriptors.
\newblock IEEE Transactions on Pattern Analysis and Machine Intelligence
  \textbf{33} (2011)  43--57

\bibitem{daisy}
Tola, E., Lepetit, V., Fua, P.:
\newblock A fast local descriptor for dense matching.
\newblock (2008)

\bibitem{Ke04pca-sift:a}
Ke, Y., Sukthankar, R.:
\newblock Pca-sift: A more distinctive representation for local image
  descriptors.
\newblock (2004)  506--513

\bibitem{7298850}
Balntas, V., Tang, L., Mikolajczyk, K.:
\newblock Bold - binary online learned descriptor for efficient image matching.
\newblock In: 2015 IEEE Conference on Computer Vision and Pattern Recognition
  (CVPR). (2015)  2367--2375

\bibitem{NIPS2014_5349}
Zhou, B., Lapedriza, A., Xiao, J., Torralba, A., Oliva, A.:
\newblock Learning deep features for scene recognition using places database.
\newblock In Ghahramani, Z., Welling, M., Cortes, C., Lawrence, N.D.,
  Weinberger, K.Q., eds.: Advances in Neural Information Processing Systems 27.
\newblock Curran Associates, Inc. (2014)  487--495

\bibitem{6909640}
Sun, Y., Wang, X., Tang, X.:
\newblock Deep learning face representation from predicting 10,000 classes.
\newblock In: 2014 IEEE Conference on Computer Vision and Pattern Recognition.
  (2014)  1891--1898

\bibitem{1640964}
Hadsell, R., Chopra, S., LeCun, Y.:
\newblock Dimensionality reduction by learning an invariant mapping.
\newblock In: 2006 IEEE Computer Society Conference on Computer Vision and
  Pattern Recognition (CVPR'06). Volume~2. (2006)  1735--1742

\bibitem{DBLP:journals/corr/VariorHW16}
Varior, R.R., Haloi, M., Wang, G.:
\newblock Gated siamese convolutional neural network architecture for human
  re-identification.
\newblock CoRR \textbf{abs/1607.08378} (2016)

\bibitem{DBLP:journals/corr/LinMCVG15}
Lin, J., Mor{\`{e}}re, O., Chandrasekhar, V., Veillard, A., Goh, H.:
\newblock Deephash: Getting regularization, depth and fine-tuning right.
\newblock CoRR \textbf{abs/1501.04711} (2015)

\bibitem{Chechik:2010:LSO:1756006.1756042}
Chechik, G., Sharma, V., Shalit, U., Bengio, S.:
\newblock Large scale online learning of image similarity through ranking.
\newblock J. Mach. Learn. Res. \textbf{11} (2010)  1109--1135

\bibitem{NIPS2016_6464}
Ustinova, E., Lempitsky, V.:
\newblock Learning deep embeddings with histogram loss.
\newblock In Lee, D.D., Sugiyama, M., Luxburg, U.V., Guyon, I., Garnett, R.,
  eds.: Advances in Neural Information Processing Systems 29.
\newblock Curran Associates, Inc. (2016)  4170--4178

\bibitem{Yu2010RelaxedCA}
Yu, Y., Yang, M., Xu, L., White, M., Schuurmans, D.:
\newblock Relaxed clipping: A global training method for robust regression and
  classification.
\newblock In: NIPS. (2010)

\bibitem{DBLP:journals/corr/WuMSK17}
Wu, C., Manmatha, R., Smola, A.J., Kr{\"{a}}henb{\"{u}}hl, P.:
\newblock Sampling matters in deep embedding learning.
\newblock CoRR \textbf{abs/1706.07567} (2017)

\bibitem{cosface}
Wang, H., Wang, Y., Zhou, Z., Ji, X., Li, Z., Gong, D., Zhou, J., Liu, W.:
\newblock Cosface: Large margin cosine loss for deep face recognition.
\newblock (2018)

\bibitem{DBLP:journals/corr/BalntasLVM17}
Balntas, V., Lenc, K., Vedaldi, A., Mikolajczyk, K.:
\newblock Hpatches: {A} benchmark and evaluation of handcrafted and learned
  local descriptors.
\newblock CoRR \textbf{abs/1704.05939} (2017)

\bibitem{DBLP:journals/corr/MishkinMPL15}
Mishkin, D., Matas, J., Perdoch, M., Lenc, K.:
\newblock Wxbs: Wide baseline stereo generalizations.
\newblock CoRR \textbf{abs/1504.06603} (2015)

\end{thebibliography}

\end{document}